\documentclass[letterpaper]{article} %
\usepackage{aaai2026}  %
\usepackage{times}  %
\usepackage{helvet}  %
\usepackage{courier}  %
\usepackage[hyphens]{url}  %
\usepackage{graphicx} %
\usepackage{natbib}  %
\usepackage{caption} %
\usepackage{algorithm}
\usepackage{algorithmic}

\usepackage{amsmath, amsfonts, amssymb, amsthm}
\usepackage{dsfont}
\usepackage{multirow}

\usepackage[scr]{rsfso}
\usepackage{makecell}
\usepackage{booktabs}
\usepackage{arydshln}

\usepackage{newfloat}
\usepackage{listings}
\DeclareCaptionStyle{ruled}{labelfont=normalfont,labelsep=colon,strut=off} %
\floatstyle{ruled}
\newfloat{listing}{tb}{lst}{}
\floatname{listing}{Listing}
\title{Decoding-Free Sampling Strategies for LLM Marginalization}
\author{
    David Pohl\textsuperscript{\rm 1}\equalcontrib\thanks{Correspondence to david.pohl@nlp.c.titech.ac.jp}, Marco Cognetta\textsuperscript{\rm 1}\equalcontrib, Junyoung Lee\textsuperscript{\rm 2}, Naoaki Okazaki\textsuperscript{\rm 1}\\
}
\affiliations{
    \textsuperscript{\rm 1}Institute of Science Tokyo, Japan\\
    \textsuperscript{\rm 2}Independent Researcher
}

\usepackage{bibentry}

\begin{document}

\maketitle

\begin{abstract}
Modern language models operate on subword-tokenized text in order to make a trade-off between model size, inference speed, and vocabulary coverage. A side effect of this is that, during inference, models are evaluated by measuring the probability of only the specific tokenization produced as the output, despite there being many possible ways to represent the same text with a subword vocabulary. Recent studies have argued instead for evaluating LLMs by \textit{marginalization} --- the probability mass of all tokenizations of a given text.

Marginalization is difficult due to the number of possible tokenizations of a text, so often approximate marginalization is done via sampling. However, a downside of sampling is that an expensive generation step must be performed by the LLM for each sample, which limits the number of samples that can be acquired given a runtime budget, and therefore also the accuracy of the approximation. Since computing the probability of a sequence given the tokenization is relatively cheap compared to actually generating it, we investigate sampling strategies that are \textit{decoding-free} --- they require no generation from the LLM, instead relying entirely on extremely cheap sampling strategies that are model and tokenizer agnostic.

We investigate the approximation quality and speed of decoding-free sampling strategies for a number of open models to find that they provide sufficiently accurate marginal estimates at a small fraction of the runtime cost and demonstrate its use on a set of downstream inference tasks.
\end{abstract}

\section{Introduction}

\begin{figure}
    \centering
    \includegraphics[width=.99\linewidth]{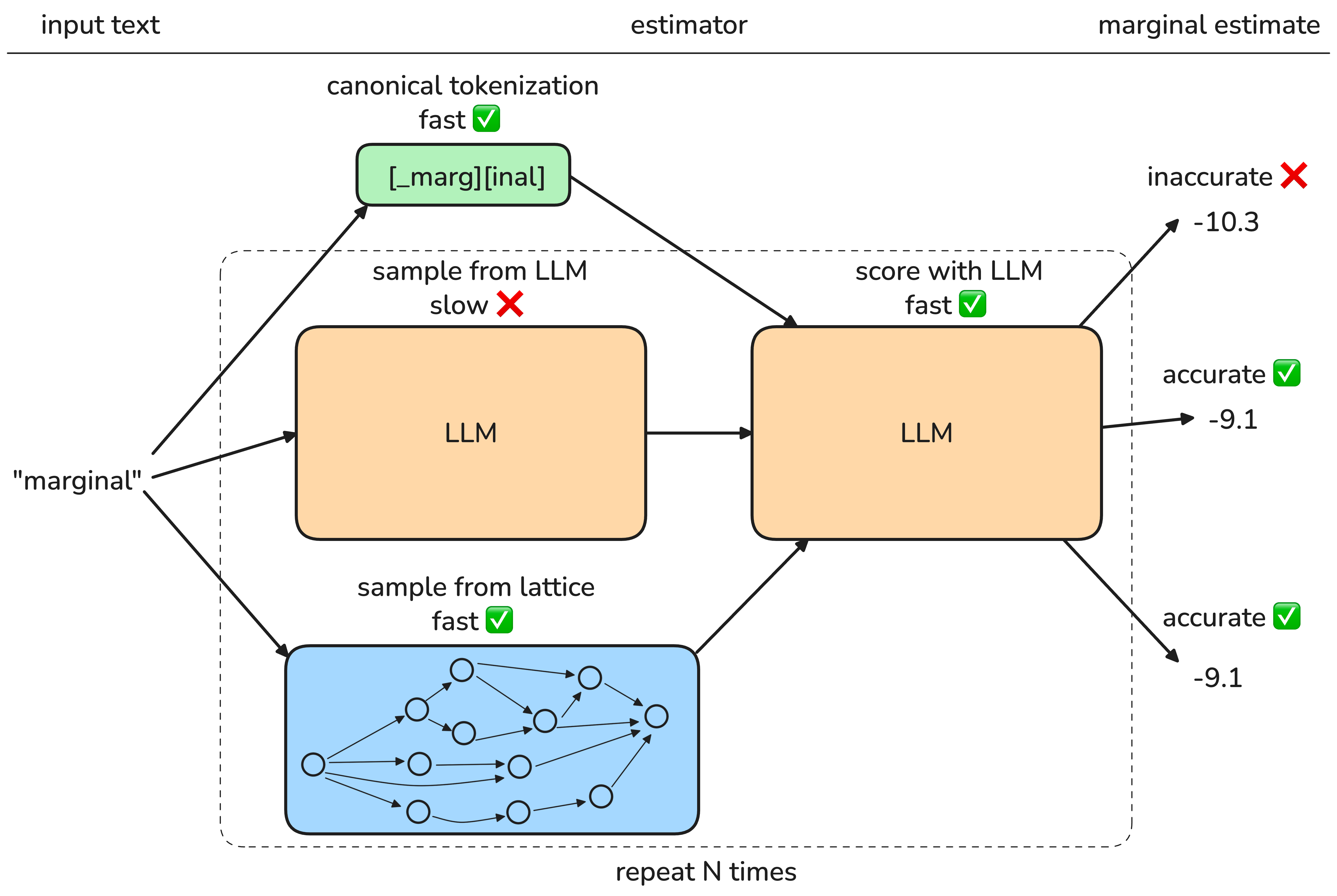}
    \caption{Three methods for approximating the marginal of an input text. The typical approach is to compute the probability of the canonical tokenization, which is fast but inaccurate. Importance sampling allows for more accurate marginalization, but sampling from an LLM is very slow. Our lattice approach is fast and produces good estimates.}
    \label{fig:system_overview}
\end{figure}

Language models consume and generate text at a subword granularity. In particular, when generating text, they output a subword sequence that is detokenized to give the raw output text. However, this subword-sequence-to-text mapping is many-to-one, as there are an exponential number of possible ways to tokenize an input sequence under a given subword vocabulary. A common practice is to  consider the \textit{canonical} tokenization of a text (that is, the subword sequence produced by the model's tokenizer when given the input) or the single sequence that is generated by the model (under some decoding policy that is designed to find the maximum probability sequence given the model and context) as the representative tokenization of an input to be used a proxy for the probability of its detokenized surface form. Formally, given a (text) sequence $\mathbf{s}$, we often approximate the probability \[ p(\mathbf{s}) \approx p(t, \mathbf{s}) = \prod_{i}^{|t|}P_{\theta}(t_i \mid t_{<i}),\] where $t$ is some valid tokenization of $\mathbf{s}$, and $P_{\theta}(t_i \mid t_{<i})$ is a model's auto-regressive next token probability according to the parameters of the model, $\theta$. However, an argument can be made that we should instead use the \textit{marginal} probability \[p(\mathbf{s}) = \sum_{\mathbf{s} \rightsquigarrow t} p(t, \mathbf{s}),\] where $\mathbf{s} \rightsquigarrow t$ is the set of all possible tokenizations of $\mathbf{s}$.

The marginal is difficult to compute, since $\mathbf{s} \rightsquigarrow t$ can be very large (exponential in the length of $\mathbf{s}$ and the size of the subword vocabulary). Instead, techniques like importance sampling are often used to approximate the marginal \cite{cao-rimell-2021-evaluate, geh-etal-2024-signal}. However, importance sampling requires \textit{generating} many sequences from the underlying LLM, which is computationally expensive. We additionally find that importance sampling often significantly \textit{underestimates} the marginal across many models and tasks.

We investigate alternative marginalization strategies that are \textit{decoding-free}, in that they never require \textit{generation} by an LLM, but only \textit{scoring}, which is relatively cheap. These strategies are language model agnostic and rely only on manipulations of a \textit{subword-lattice} that succinctly encodes all possible tokenizations of an output while still remaining easy to construct and sample from. 

Several works have found that the marginal, and specifically the marginal excluding the canonical tokenization, carries meaningful signal in that it can match or outperform canonical-only inference on downstream tasks \cite{chirkova-etal-2023-marginalize,geh-etal-2024-signal,broken-tokens}. Our experimental results also support this and further motivate our contributions since we provide more accurate marginal estimates at a fraction of the inference-time cost.

\section{Related Work} \label{sec:related}

Tokenization is the first step of the modern neural language modeling pipeline, where text is converted into a subword sequence that is understandable by the model. Many algorithms exist for subword tokenization, but Byte-Pair Encoding (BPE) \cite{sennrich-etal-2016-neural} has found the most use in modern LLMs \cite{gpt, llama, gemma3, qwen}.

Tokenization is generally a deterministic process, but, given a subword vocabulary, there are often many ways to represent a given input text and recent research has considered the problem of computing the \textit{marginal} of an input --- the total probability mass assigned to all possible tokenizations of a given input sequence, not just the canonical tokenization from the tokenizer \cite{cao-rimell-2021-evaluate,chirkova-etal-2023-marginalize,geh-etal-2024-signal}.

In particular, \citet{cao-rimell-2021-evaluate} find that marginalization helps improve LM accuracy, especially on out-of-domain data, and \citet{geh-etal-2024-signal} find using the marginal of \textit{non-canonical} tokenizations can even out-perform traditional decoding methods on Q\&A tasks. Both of these use importance sampling, a method for estimating difficult-to-sample-from distributions, for estimating the marginal. This is necessary as computing the marginal is known to be NP-Hard in general \cite{geh-etal-2024-signal}.

Other work has analyzed and improved language models' ability to handle non-canonical tokenization by replacing static tokenizers with stochastic tokenizers so that models are exposed to many different tokenizations of the same text \cite{kudo-2018-subword, provilkov-etal-2020-bpe, cognetta-etal-2024-distributional, bauwens-etal-2025-grampa}. Additionally, \citet{broken-tokens} found that, even LLMs that are trained with deterministic tokenizers are robust to major perturbations of the input tokenization. Further, several works have found that LLMs have implicit knowledge of their subword vocabulary surface forms, even without being exposed to them directly \cite{kaushal-mahowald-2022-tokens, DBLP:journals/corr/abs-2402-09808, edman-etal-2024-cute}, again implying that non-canonical tokenizations carry meaningful information.

\section{Background and Notation} \label{sec:background}

\subsection{Subword Tokenization}
\textit{Subword tokenization} represents raw text at an intermediate granularity between characters and words \cite{mielke2021wordscharactersbriefhistory}. The basis of subword tokenization is a \textit{subword vocabulary}, the set of subword tokens that can be used for representation. Given a character alphabet $\Sigma$, a subword vocabulary $\Gamma$ is a finite set of strings made up of characters from $\Sigma$, with the additional requirement that $\Sigma \subseteq \Gamma$.

\textit{Subword tokenizers} convert character sequences over $\Sigma^*$ to subword sequences over $\Gamma^*$. Typically, subword tokenizers are deterministic (each possible input is mapped to a single, canonical output). We write $T: \Sigma^* \rightarrow \Gamma^*$ for the tokenization process and $T^{-1} : \Gamma^* \rightarrow \Sigma^*$ for the inverse detokenization process. Unlike the tokenization process, the detokenization process is not one-to-one. For example, a tokenizer might always produce $T(\texttt{sampler}) = \texttt{[\_sam]\![pler]}$ but the detokenization process is many-to-one: $T^{-1}(\texttt{[\_sa]\![mp][ler]}) = T^{-1}(\texttt{[\_s]\![ample]\![r]}) = \dots = T^{-1}(\texttt{[\_s]\![a]\![m]\![p]\![l]\![e]\![r]}) = \texttt{sampler}$.

Subword tokenizers can also be \textit{stochastic} by injecting randomness into the tokenization process to produce a probability distribution of sequences \cite{kudo-2018-subword, provilkov-etal-2020-bpe}.

\subsection{Building a Tokenization Lattice} \label{ssec:tokenization_lattice}

\begin{figure}
    \centering
    \includegraphics[width=\linewidth]{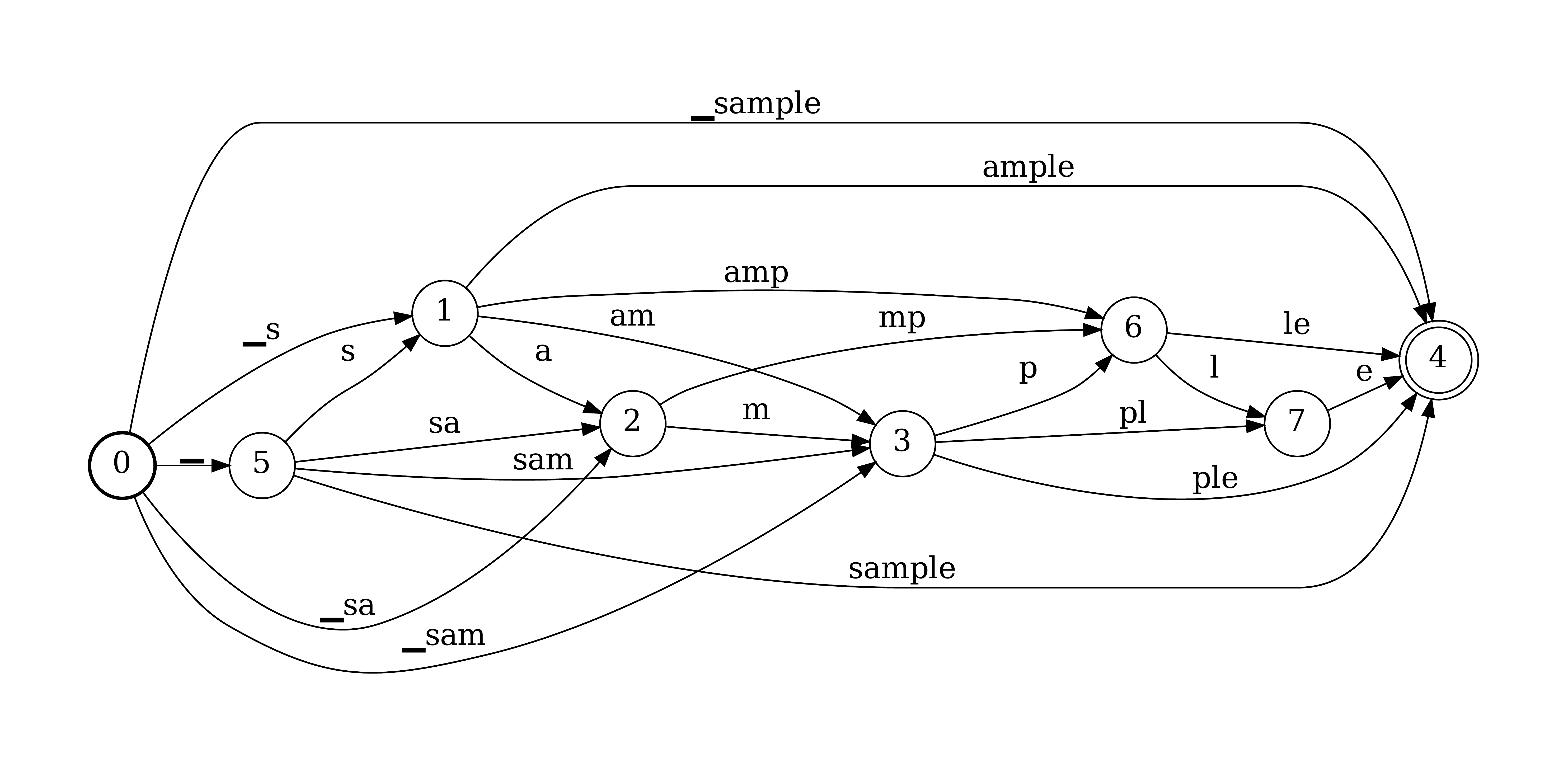}
    \caption{An example subword tokenization lattice for the input $\mathbf{s} = \texttt{sample}$ using the \textsc{Llama2} vocabulary.}
    \label{fig:subword_lattice}
\end{figure}

We utilize finite automata and transducers to build a \textit{lattice} of subword tokenizations for an input --- an automaton that encodes all subword sequences that, when detokenized, would match the input \cite{willard2023efficientguidedgenerationlarge,cognetta2024tokenizationfinitestatetransduction,koo}. An example is shown in Figure \ref{fig:subword_lattice}. We only briefly introduce the necessary concepts here, and point the reader to \citet{allauzen-mohri-2009} for a more detailed overview of automata and transducer algorithms.

 Suppose we have a sequence $\mathbf{s}$ and a subword vocabulary $\Gamma$ (irrespective of the underlying tokenization algorithm). We wish to find the full set of tokenizations $t$ which corresponds to $\mathbf{s}$. Let $\mathcal{T}_{\Gamma}$ be a character-to-subword transducer $\mathcal{T}_{\Gamma}: \Sigma^* \rightarrow \Gamma$ which recognizes the relation $\bigcup_{t \in \Gamma} (T^{-1}(t), t)$. In words, this maps sequences of characters in $\Sigma$ to a subword token $t \in \Gamma$. After closure, we have a transducer $\mathcal{T}_{\Gamma}^*: \Sigma^* \rightarrow \Gamma^*$ which maps sequences of characters to \textit{sequences} of subwords in $\Gamma$. Then, given $\mathbf{s} \in \Sigma^*$, we represent it as an automaton (which only accepts $\mathbf{s}$) $\mathcal{A}_{\mathbf{s}}$. 

 Let $\mathcal{A} \circ\mathcal{T}$ be the \textit{composition} operator between an automaton and a transducer. This operator computes the relation \[ \mathcal{A} \circ \mathcal{T} = \{(\mathbf{x}, \mathbf{y}) \mid \mathbf{x} \in \mathcal{L}(\mathcal{A}) \wedge (\mathbf{x}, \mathbf{y}) \in \mathcal{L}(\mathcal{T})\},\]
the subset of transductions encoded by $\mathcal{T}$ where the input is in the language of $\mathcal{A}$.

Additionally, let $\textsc{Proj}: \Sigma^* \times \Gamma^* \rightarrow \Gamma^*$ be a unary \textit{projection} operator which transforms a transducer to an automaton by computing the set $\textsc{Proj}(\mathcal{T}) = \{\mathbf{y} \mid (\mathbf{x}, \mathbf{y}) \in \mathcal{L}(\mathcal{T})\}$ of sequences in $\Gamma^*$ such that there is some input sequence in $\Sigma^*$ that $\mathcal{T}$ transduces to it.

Then, computing $\textsc{Proj}(\mathcal{A}_{\mathbf{s}} \circ \mathcal{T}_{\Gamma}^*)$ produces the subword lattice of $\mathbf{s}$.

\subsection{Sampling from LLMs}

Sampling sequences from an LLM involves repeatedly sampling tokens from the model's internal autoregressive probability distribution, where a sequence $t_1t_2\dots t_n \in \Gamma^*$ is sampled with probability:
\[p(t_1t_2\dots t_n) = \prod_{i=1}^{|t|} P_{\theta}(t_i \mid t_{<i}),\]
where $P_{\theta}(t_i \mid t_{<i})$ is the probability distribution over next tokens, given a context $t_1t_2\dots t_{i-1}$, produced by the softmax output layer of a language model parameterized by $\theta$.

After each sample, the sampled token is added to the context and the process is repeated until we sample the end-of-sequence token or reach a maximum length.

\subsection{Constrained Sampling}

We often wish to constrain an LLM with vocabulary $\Gamma$ to generate only sequences which match a given pattern. Let $\mathcal{A}^{sub}$ be a subword-level pattern constructed by promoting a character-level pattern $\mathcal{A}$ via $\textsc{Proj}(\mathcal{A} \circ \mathcal{T}_{\Gamma}^*)$.

We wish to modify the sequential probability distribution of the LLM so that $p'(t_1t_2\dots t_n) = 0$ if $t_1t_2\dots t_n \notin \mathcal{L}(\mathcal{A}^{sub})$. A common way to do this is to \textit{locally-constrain} the autoregressive probability computation:
\[P'_{\theta}(t_i \mid t_{<i}, \mathcal{A}^{sub}) = \begin{cases}
    \frac{P_{\theta}(t_i \mid t_{<i})}{z}, & \text{if } t_1t_2\dots t_i\Gamma^* \subseteq \mathcal{A}^{sub} \\
    0 & \text{otherwise.}
\end{cases}\]

When generating a token $t_i$ to append to a context $t_{<i}$, $P'_{\theta}$ assigns probability 0 to all tokens in $\Gamma$ that would prevent the sequence from matching the underlying pattern and uses the original autoregressive probability, scaled by a normalization factor $z$, for all valid tokens.

Then, our constrained distribution is
\[p'(t_1t_2\dots t_n, \mathcal{A}^{sub}) = \prod_{i=1}^{|t|} P'_{\theta}(t_i \mid t_{<i}, \mathcal{A}^{sub}),\] which, by the product chain rule, assigns probability 0 to all subword sequences not in $\mathcal{A}^{sub}$, as desired.

We can use this to sample only tokenizations of an input sequence $\mathbf{s}$ by using $\mathcal{A}_{\mathbf{s}}^{sub} = \textsc{Proj}(\mathcal{A}_{\mathbf{s}} \circ \mathcal{T}_{\Gamma}^*)$, which will be useful in the remainder of this work.

\section{Marginalization}

Marginalization computes the probability of the \textit{surface form} of $\mathbf{s}$ instead of just a single tokenization of $\mathbf{s}$. In general, marginalization is intractable, since there are an exponential number of possible tokenizations of an input $\textbf{s}$, but approximate marginalization algorithms exist. For language models, these algorithms typically take samples from the model's distribution (or a related one) and use the sample probabilities to estimate the total probability mass \cite{cao-rimell-2021-evaluate, geh-etal-2024-signal, vieira2025languagemodelscanonicalbytepair}.

The simplest is \textit{rejection sampling}, where we sample directly from $p$, but ``reject" samples that do not match our target surface form. But, an issue with rejection sampling is that if $p(\mathbf{s})$ is small, we may only rarely sample valid tokenizations, and so this sum will take a long time to converge.

Instead, we can use importance sampling, where we define a \textit{proxy} distribution $q$ (as opposed to the \textit{primary} distribution, $p$) which only produces samples that match $\mathbf{s}$:
\[p(\mathbf{s}) = \mathbb{E}_{t \sim q}\left[\frac{p(t)}{q(t)}\right] \approx \frac{ 1 }{N}\sum_{i}^N \frac{p(t^{(i)})}{q(t^{(i)})}. \]

In line with prior work, we use the locally-masked distribution for constrained LLM next-token prediction based on the subword lattice \cite{geh-etal-2024-signal}. Let $\mathbf{s}$ be our intended surface form and $\mathcal{A}_{\mathbf{s}}^{sub}$ be the lattice of all tokenizations of $\mathbf{s}$. Then, we set our proxy distribution to be $p'(\boldsymbol{\cdot}, \mathcal{A}_{\mathbf{s}}^{sub})$, which allows us to only generate sequences which match $\mathbf{s}$.

\subsection{Full and Partial Enumeration} \label{ssec:enumeration}
In principle, we can exactly compute the marginal by enumerating all sequences in $\mathcal{A}^{sub}_{\textbf{s}}$. If $\mathbf{s}$ is short, this may be possible, but as $\mathbf{s}$ grows it becomes infeasible due to the number of tokenizations of a sequence being exponential in the length of $\mathbf{s}$. Additionally, \citet{geh-etal-2024-signal} showed that computing the marginal exactly is $\textsc{NP}$-Hard for LLMs.

However, note that \textit{any} partial enumeration (i.e., a set of sequences with no duplicates) acts a lower-bound on the true marginal which monotonically increases towards the true bound as you increase the number of samples.

\section{The Distribution of Probability Mass} \label{sec:probability_mass_distribution} %
Language models are trained on the canonical tokenizations of text, which causes the bulk of the probability mass for a given surface form to be held by its canonical tokenization. In practice, this means that the canonical tokenization is a reasonable approximation for the marginal. However, the non-canonical marginal has been shown to carry significant signal as well \cite{geh-etal-2024-signal}. Therefore, we investigate approximating the non-canonical marginal.

 Excluding the canonical tokenization from marginal computation offers the following two key benefits. First, for sampling schemes, a large number of samples will correspond to the canonical tokenization, meaning we are not able to explore much of the tokenization space and we see a less diverse set of samples. And second, an estimate of the non-canonical marginal can be turned into an estimate of the full marginal by simply adding the probability of the canonical sequence to it.

\subsubsection{A Lot of Mass is ``Near" the Canonical Sequence}

Empirically, we observed that most of the marginal is contained in the canonical sequence like in \cite{geh-etal-2024-signal}. But, if we consider only the non-canonical marginal, we observed in our experiments that a huge portion of the probability mass is contained in sequences that are \textit{almost} the canonical sequence. For example, sequences which are identical to the canonical except one token is split into two. 

Since many tokenization algorithms have their roots in compression, it is unsurprising that canonical tokenizations tend to have nearly the optimal sequence length for an input sequence and vocabulary. It has been shown that language models have a bias towards shorter sequences due to the autoregressive decoding strategy \cite{murray-chiang-2018-correcting, DBLP:journals/tmlr/WelleckBFSXNKH24}. Intuitively then, it is not surprising that other short sequences would have relatively high probability, especially if they are very similar to the canonical sequence except for a few tokens \cite{chirkova-etal-2023-marginalize}.

We will exploit this in Section \ref{sec:out_estimation_algorithm} to produce a fast estimate of the marginal by lower-bounding it with the probability of tokenizations that are close to the canonical.

\subsection{Downsides of a Constrained LLM as a Proxy}

\begin{figure}
    \centering
    \includegraphics[width=\linewidth]{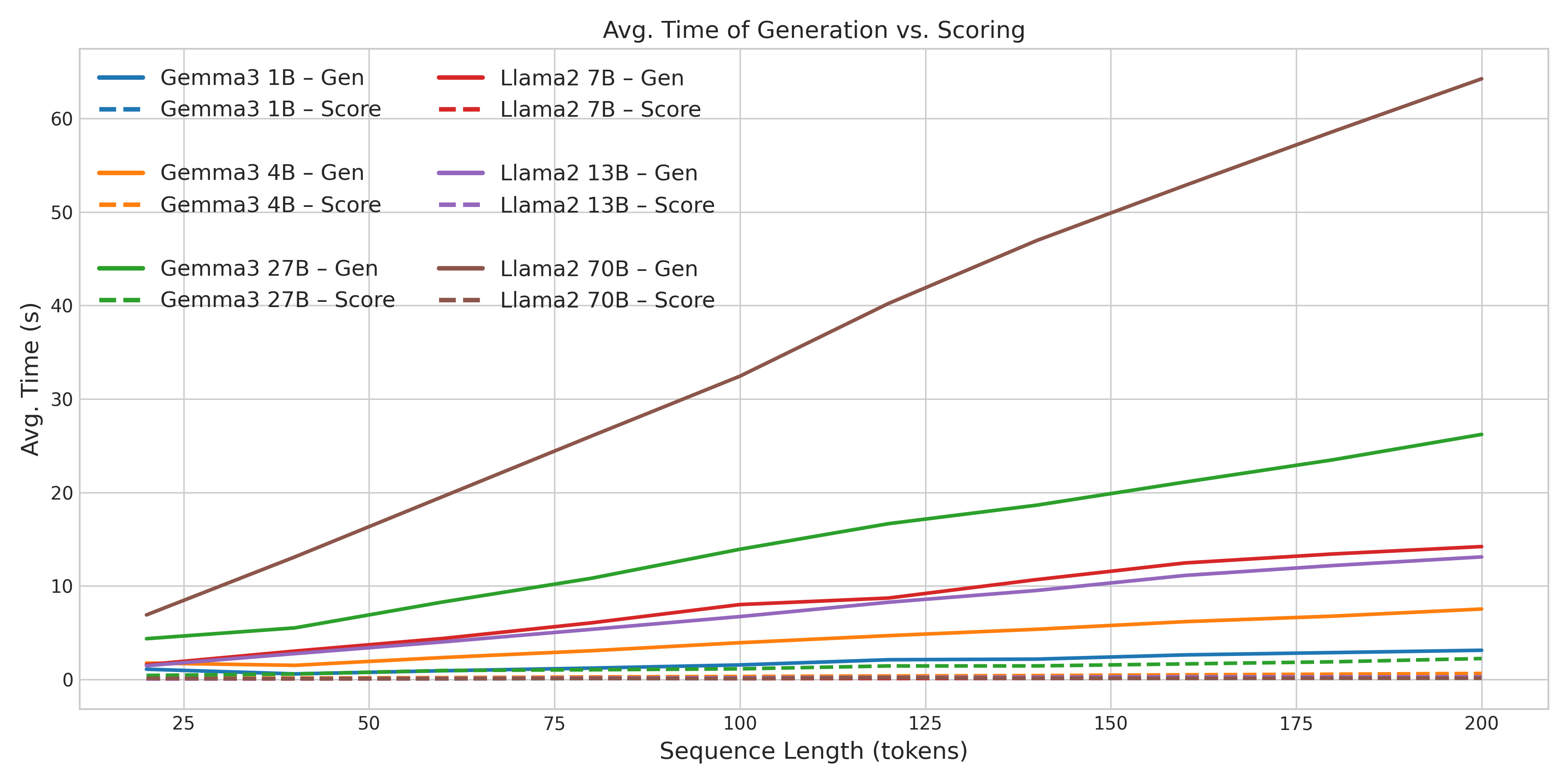}
    \caption{The runtimes of \textit{generating} a sequence from an LLM and \textit{scoring} a sequence with an LLM for \textsc{Llama} and \textsc{Gemma}. Note that, as the sequence length grows, generation becomes significantly slower than scoring.}
    \label{fig:timing}
\end{figure}

There are two primary downsides to using a proxy distribution based on the locally-masked distribution $P'_{\theta}$: speed and sample quality. 

Sampling sequences in language models (constrained or unconstrained)  is computationally expensive, especially compared to scoring an existing subword sequence. Figure \ref{fig:timing} shows the wall-clock runtime penalty of sampling from the LLM due to the quadratic runtime of self attention.

The slowdown is largely caused by generation requiring $O(n)$ calls to the forward pass\footnote{We focus on forward passes rather than asymptotic runtime complexity to abstract away details of the architecture and the use of techniques like KV-caching.} of a model to generate a sequence of length $n$, which must be performed serially. On the other hand, scoring a sequence requires only one forward pass which can be parallelized across the sequence length.

The second issue is that the local-masking distribution $P'_{\theta}$ produces a warped autoregressive distribution that is not necessarily representative of the original, unconstrained language model's distribution \cite{vieira2025languagemodelscanonicalbytepair}. Table \ref{tab:warped_inversions} contains Spearman $\rho$ values for the relative orderings of a language model and it's constrained version for a series of surface forms. The warped distribution is only moderately correlated with the original distribution (that is, not only does $p(t_1) \ne p'(t_1)$, but also $p(t_1) > p(t_2)$ does not necessarily imply $p'(t_1) > p'(t_2)$). 

\begin{table}[]
\centering
\begin{tabular}{lccc}
\toprule
\multirow{2}{*}{\textsc{Model}} & \multicolumn{3}{c}{\textsc{Spearman's $\rho$}}      \\
      & $|\mathbf{s}|\approx50$ & $|\mathbf{s}|\approx100$       & $|\mathbf{s}|\approx300$ \\ \midrule
\textsc{Llama2-7b} & $0.70 \pm 0.16$    & $0.66 \pm 0.10$          & $0.75 \pm 0.11$     \\ \midrule
\textsc{Llama2-13b} & $0.64 \pm 0.15$    & $0.68 \pm 0.11$          & $0.76 \pm 0.09$     \\ \midrule
\textsc{Gemma3-1b} & $0.69 \pm 0.17$    & $0.63 \pm 0.14$          & $0.65 \pm 0.12$     \\ \midrule
\textsc{Gemma3-4b} & $0.71 \pm 0.19$    & $0.58 \pm 0.13$          & $0.58 \pm 0.11$     \\ \midrule
\textsc{Gemma3-12b} & $0.80 \pm 0.13$    & $0.70 \pm 0.12$          & $0.64 \pm 0.09$     \\ \bottomrule
\end{tabular}
\caption{Spearman's $\rho$ for the rankings of sequences by probability under $p$ (base language model) and $q$ (the constrained language model) at different sequence lengths. Each sequence length test-set contains 15 sequences. For each sequence, we generate 1000 unique tokenizations by sampling from a constrained language model and then compare the sequences' relative rankings. There is only moderate positive correlation between the rankings from $p$ and $q$, which can cause importance sampling from $q$ to converge slowly.
}\label{tab:warped_inversions}
\end{table}

In Section \ref{ssec:underestimation_experiment}, we also show that importance sampling tends to severely underestimate the marginal, possibly due to over-sampling low-probability sequences due to the imperfect alignment of the proxy and primary distributions' orderings.

\section{Generating Tokenizations Efficiently}

Rather than sampling from the constrained LLM, we aim to generate tokenizations in a different and cheaper way by utilizing the tokenization lattice from Section \ref{ssec:tokenization_lattice}.

Here, we discuss only the sampling methods used in our final implementation. There are a variety of stochastic tokenization methods that could be used like BPE-Dropout \cite{provilkov-etal-2020-bpe}, MaxMatch-Dropout \cite{hiraoka-2022-maxmatch}, UnigramLM \cite{kudo-2018-subword}, and other variants \cite{hiraoka-etal-2019-stochastic, broken-tokens}. However, each of these were insufficient in some way: they do not allow for controlling tokenization output length, they cannot be sampled without replacement, they require a training step, or they are dependent on a specific underlying tokenization algorithm, so we do not cover them here.

\subsection{Uniformly Sampling Tokenizations} \label{ssec:uniform_sampling}

Let $\mathbf{s} \in \Sigma^*$ be the input text and let $ \mathcal{A}^{sub}_{\mathbf{s}} = \textsc{Proj}(\mathcal{A}_{\mathbf{s}} \circ \mathcal{T}_{\Gamma}^*)$ be the tokenization lattice, which is a DAG where every path describes a valid tokenization of $\mathbf{s}$ according to the tokeinzer's vocabulary $\Gamma$. Suppose there are $N$ possible tokenizations, then sampling uniformly would mean sampling each individual path with probability $\frac{1}{N}$.

We briefly describe a linear time algorithm to sample uniformly from a DAG, and point the reader to \cite{hoens2008counting, bauwens-etal-2025-grampa} for a more thorough treatment of uniformly sampling tokenizations.

Let $D$ be a DAG with $v_s$ and $v_f$ being the start and final states, respectively, and let $n(u)$ be the neighbors of a state $u$. Then we define a recurrence as $T_{v_f} = 1$ as the base-case (since it has no outgoing arcs) and $T_{u} = \sum_{v \in n(u)} T_{v}$ for the general case. Then, by computing $T_{v_s}$ (the total number of paths in the DAG), we also know each $T_u$. Notice that, if we interpret a given $u$ as the start state, then the set of reachable states from $u$ also forms a DAG.

Now, we associate each of the $N = T_{v_s}$ distinct paths with an integer from 1 to $N$. Suppose we sample a value $z \sim  \mathcal{U}(1, N)$ (the uniform distribution of integers between 1 and $N$, inclusive). Starting from the start state $v_s$, iterate over $n(v_s)$ in some predefined order and check which state must the $z$\textsuperscript{th} path pass through. We then travel to that state, and adjust the current path index (by subtracting the number of paths that lead to neighboring states that we checked before the state we ended up traveling to). We recurse on this until we reach the final state $v_f$ and return the subword sequence corresponding to the sequence of arcs that we traversed.

Each local sample from state $u$ can be done in constant time since the maximum out-degree at a state is equal to the length of the longest token in the vocabulary, which we consider to be a constant. The total runtime for a single sample is therefore $O(n)$.

We can use this to sample paths uniformly without replacement by instead sampling without replacement from $\mathcal{U}(1, N)$ and then extracting each path.

\subsection{Length-Constrained Sampling}

Uniformly from the set of all tokenizations tends to produce very long tokenizations (that is, the final tokenization tends to consist primarily of short tokens). On the other hand, most tokenizers are trained with the implicit objective of \textit{compressing} the sequence length \cite{galle-2019-investigating}. Thus, we hypothesize that a perfectly uniform sampling will produce primarily low-probability sequences; a thought that has been considered before as well by \citet{bauwens-etal-2025-grampa}, who propose a similar, but biased sampling procedure designed to induce shorter tokenizations which should be more meaningful.

Instead, based on our empirical findings from Section \ref{sec:probability_mass_distribution}, we wish to simply restrict the samples to only those with a specified maximum length. Consider an automaton that accepts the language $\Gamma^{\le\ell}$, the set of all subword sequences of length at most $\ell$ made up of tokens from $\Gamma$. Then $\mathcal{A}_{\mathbf{s}}^{sub} \cap \Gamma^{\le \ell}$ is another tokenization lattice which only accepts sequences of length at most $\ell$. Since this automaton is also a DAG, we can sample uniformly from it exactly as in Section \ref{ssec:uniform_sampling}.

\subsection{Enumerating Around the Canonical Sequence}

As described in Section \ref{sec:probability_mass_distribution}, we expect that much of the non-canonical tokenization mass is carried by tokenizations that are close to the canonical tokenization. We can easily enumerate a subset of close-to-canonical tokenizations. Let $t_1t_2\dots t_n$ be the canonical tokenization sequence and let $\textsc{decompose}(t)$ be the set of tokenization sequences $t'_1t'_2 \in \Gamma^2$ such that $T^{-1}(t) = T^{-1}(t'_1t'_2)$. Then, we can form the full set of ``off-by-one" tokenizations by building the set of sequences where we select one token $t_i$ from the canonical sequence and replace it with one of the sequences from $\textsc{decompose}(t_i)$. Let $m$ be the length of the longest token in $\Gamma$. Then, we can fully enumerate this set of ``off-by-one" tokenizations, as it contains only $O(mn)$ distinct sequences. Since $m$ is often very small (e.g., for \textsc{Llama2}, $m = 16$), this enumeration is cheap.

\section{Our Marginal Estimation Algorithm} \label{sec:out_estimation_algorithm}

We now have all the components for our marginal estimation algorithm. Instead of using importance sampling, we enumerate sequences that are likely to be high-probability, and then sample uniformly without replacement from the rest. This produces a lower bound on the marginal that monotonically improves with an increased sample size.

Algorithm \ref{alg:our_algorithm} contains our lattice sampler pseudocode. We produce a sample of non-canonical subword sequences that are to be scored by the language model. We seed the list with the set of ``off-by-one" tokenizations before uniformly sampling (without replacement) sequences of length at most $\ell$ from the remaining lattice. This results in a large set of unique sequences that are likely to have high probability.

We score all sequences at once using the language model and report the output as our marginal probability estimate. Each of the steps in Algorithm \ref{alg:our_algorithm} are computationally inexpensive (especially relative to the time it takes to generate a sequence with an LLM) and scoring them is also fast (Table \ref{fig:timing}), so overall this sampling and scoring procedure has a significant runtime advantage over traditional proxy importance sampling.

As an example, in our implementation sampling $N=10000$ paths without replacement from a lattice for a sentence with 300 letters and the \textsc{Gemma3} tokenizer takes slightly less than 1 second on a consumer CPU. Thus, this time is negligible compared to the time required to generate the same number of sequences using LLM decoding.

\begin{algorithm}[tb]
\caption{Our Lattice Sampler}
\label{alg:our_algorithm}
\textbf{Input}: Surface Form $\mathbf{s}$, Vocabulary $\Gamma$, Samples $k$, Max Length $\ell$ \\
\textbf{Returns}: $k$ distinct sequences from $\Gamma^*$ \\
\begin{algorithmic}[1]
\STATE Let $t$ be the canonical tokenization of $\mathbf{s}$
\STATE Let $\mathcal{O}_{\mathbf{s}}$ be the set of off-by-one tokenizations of $t$
\STATE Let $\mathcal{A}_{\mathbf{s}}^{sub} = \textsc{Proj}(\mathcal{A}_{s} \circ \mathcal{T}_{\Gamma}^*)$ be the lattice for $\mathbf{s}$
\STATE Compute $\mathcal{A}_{final} = \mathcal{A}_{\mathbf{s}}^{sub} \cap \overline{\mathcal{O}_{\mathbf{s}}} \cap \Gamma^{\le \ell}$
\STATE Let $N$ be the number of paths in $\mathcal{A}_{final}$
\IF {$k - |\mathcal{O}_{\mathbf{s}}| > 0$}
\STATE Let $I \sim \mathcal{U}(1, N)$ be a set of $k - |\mathcal{O}_{\mathbf{s}}|$ distinct values
\FOR {$i \in I$}
    \STATE Add path $\pi_i$ from $\mathcal{A}_{final}$ to $\mathcal{O}_{\mathbf{s}}$
\ENDFOR
\ENDIF
\STATE \textbf{return} $\mathcal{O}_{\mathbf{s}}$
\end{algorithmic}
\end{algorithm}

\section{Experiments}

We conducted two downstream experiments, Q\&A and translation, to determine if 1) non-canonical marginalization is an effective inference strategy and 2) our method produces better marginal estimates with less time. For each experiment setting, we used a variety of language models (the \textsc{Llama2} \cite{llama} and \textsc{Gemma3} \cite{gemma3} families with various parameter counts). We used the same prompts and generation hyperparameters (see supplementary material) and only varied the marginalization algorithm. All of our models used the base \textsc{HuggingFace} inference engine \cite{wolf-etal-2020-transformers} with a custom constrained logit manipulator layer on top of it for importance sampling. We used a cluster of \textsc{NVIDIA H100} GPUs for inference (we used a single GPU when possible for small models).

To isolate only the effect of the marginal and the marginalization algorithm, we did not use any decoding strategies like RAG, few-shot prompting, or chain of thought. Note that our goal is not to achieve state-of-the-art performance, but to measure the relative benefit of lattice sampling vs proxy importance sampling as alternative inference strategies to canonical tokenization scoring

\subsection{Q\&A Experiments}

\begin{table}[t]
\fontsize{9}{11}\selectfont
\centering
\begin{tabular}{@{} c@{} c@{} l c: c c c @{}}
\toprule
\multicolumn{2}{c}{\textsc{Model}} & \textsc{Dataset} & \textsc{Canon} & \textsc{Proxy} & \textsc{Lattice} & \textsc{Speedup} \\ \midrule
\multirow{6}{*}{\begin{tabular}{@{}c@{}}\rotatebox{-90}{\textsc{Llama2}}\end{tabular}} 
  & \multirow{3}{*}{\textsc{7b}} & \textsc{Book} & $49.8$ & $\mathbf{48.1}$ & $47.3$ & $3.8\times$ \\
  & & \textsc{Med} & $29.0$ & $\mathbf{31.7}$ & $31.3$ & $4.8\times$ \\
  & & \textsc{Arc} & $51.5$ & $\mathbf{49.0}$ & $47.1$ & $4.5\times$ \\
\cmidrule(lr){2-7}
  & \multirow{3}{*}{\textsc{13b}} & \textsc{Book} & $63.7$ & $\mathbf{61.6}$ & $59.6$ & $8.4\times$ \\
  & & \textsc{Med} & $36.6$ & $\mathbf{40.7}$ & $39.0$ & $8.6\times$ \\
  & & \textsc{Arc} & $68.6$ & $\mathbf{65.3}$ & $61.9$ & $5.4\times$ \\
\midrule
\multirow{9}{*}{\begin{tabular}{@{}c@{}}\rotatebox{-90}{\textsc{Gemma3}}\end{tabular}} 
  & \multirow{3}{*}{\textsc{1b}} & \textsc{Book} & $38.8$ & $32.4$ & $\mathbf{42.4}$ & $3.3\times$ \\
  & & \textsc{Med} & $32.4$ & $29.8$ & $\mathbf{31.9}$ & $2.4\times$ \\
  & & \textsc{Arc} & $47.7$ & $36.9$ & $\mathbf{38.2}$ & $1.4\times$ \\
\cmidrule(lr){2-7}
  & \multirow{3}{*}{\textsc{4b}} & \textsc{Book} & $67.8$ & $32.2$ & $\mathbf{52.9}$ & $6.2\times$ \\
  & & \textsc{Med} & $46.4$ & $29.9$ & $\mathbf{37.3}$ & $7.6\times$ \\
  & & \textsc{Arc} & $78.3$ & $39.5$ & $\mathbf{61.5}$ & $7.5\times$ \\
\cmidrule(lr){2-7}
  & \multirow{3}{*}{\textsc{27b}} & \textsc{Book} & $84.0$ & $46.0$ & $\mathbf{67.5}$ & $36.5\times$ \\
  & & \textsc{Med} & $67.9$ & $43.4$ & $\mathbf{51.9}$ & $30.2\times$ \\
  & & \textsc{Arc} & $93.4$ & $52.8$ & $\mathbf{81.1}$ & $22.9\times$ \\
\bottomrule
\end{tabular}
\caption{Q\&A accuracy results (percentage of correctly answered questions) for each model family, size, and dataset, with marginalization estimates. Lattice sampling often outperforms proxy importance sampling while using the same number of samples and substantially less inference time. \textbf{Bolded} numbers indicate which of the marginal estimators (\textsc{Proxy} vs.\ \textsc{Lattice}) had higher accuracy, since \textsc{Canon} was nearly always the best. \textsc{Speedup} is the relative speed increase for marginalization when using lattice sampling instead of importance sampling.}
\label{tab:QA_results}
\end{table}

We used three standard Q\&A datasets: \textsc{OpenBookQA}, a basic Q\&A understanding dataset \cite{OpenBookQA2018}; \textsc{MedMCQA}, a medical Q\&A dataset \cite{pmlr-v174-pal22a}; and the \textsc{AI2 Reasoning Challenge} (\textsc{Arc}) \cite{allenai:arc}. We selected a random set of 250 questions from each. For each example, we computed the probability of the canonical sequence for each answer and estimated the non-canonical marginal for each answer using proxy importance sampling and our lattice-based estimate using 1000 samples for importance sampling and 1000 tokenizations (including the set of off-by-one tokenizations) for lattice sampling, derived from Algorithm \ref{alg:our_algorithm}. We selected the answer with the highest probability and measured the accuracy of each model.

Table \ref{tab:QA_results} shows the results for each model and dataset. In nearly every setting, the canonical tokenization estimate performs the best. Curiously, both marginalization strategies beat the canonical tokenization for \textsc{MedMCQA} with both \textsc{Llama} models. Lattice sampling also slightly outperforms canonical in \textsc{OpenBook} for \textsc{Gemma3-4b}. Otherwise, canonical is by far the most accurate.

Another interesting finding is that in every case, $\textsc{Llama}$ importance sampling outperforms lattice sampling, but, conversely, in every case, \textsc{Gemma} lattice sampling outperforms importance sampling. The main difference is that the cases where importance sampling is better, the gap is small, but where lattice sampling is better, the gap is generally large.

We hypothesize that, since the answers in the dataset are very short sentences, the space of tokenizations is not that large. However, while importance sampling often resamples the same small set of tokenizations, lattice sampling is able to explore a much larger set of possible tokenizations, which produces a much more accurate estimate.

The speedup column shows that even with the relatively short sequence lengths for this dataset, lattice sampling is significantly faster (more than $30x$ for larger models) while still providing accurate marginal estimates.

These results support the claim that marginalization can carry useful signals, and strongly support
 our claim that our lattice sampler produces better marginal estimates at a fraction of the cost.

\subsection{Translation Experiments}

\begin{table}[t]
\centering
\begin{tabular}{@{} c c c c c @{}}
\toprule
\multicolumn{5}{c}{\textsc{en$\rightarrow$de}} \\
\cmidrule(lr){1-5}
\textsc{Model} & \textsc{Canon} & \textsc{Proxy} & \textsc{Lattice} & \textsc{Speedup} \\
\midrule
\textsc{Gemma3-1b}  & $\mathbf{73.0}$ & $72.7$ & $\mathbf{73.0}$ & $2.1\times$ \\
\textsc{Gemma3-4b}  & $80.8$ & $\mathbf{81.3}$ & $81.2$ & $2.9\times$ \\
\textsc{Gemma3-27b} & $84.0$ & $\mathbf{84.3}$ & $84.0$ & $9.8\times$ \\
\addlinespace[0.8ex]
\midrule
\multicolumn{5}{c}{\textsc{en$\rightarrow$cs}} \\
\cmidrule(lr){1-5}
\textsc{Model} & \textsc{Canon} & \textsc{Proxy} & \textsc{Lattice} & \textsc{Speedup} \\
\midrule
\textsc{Gemma3-1b}  & $65.8$ & $66.1$ & $\mathbf{66.8}$ & $2.2\times$ \\
\textsc{Gemma3-4b}  & $\mathbf{86.3}$ & $86.2$ & $\mathbf{86.3}$ & $3.1\times$ \\
\textsc{Gemma3-27b} & $\mathbf{88.3}$ & $88.1$ & $88.2$ & $10.0\times$ \\
\bottomrule
\end{tabular}
\caption{Translation results for our zero-shot translation pipeline. Each block is a language direction; rows are model variants, and columns are marginal approximation methods. \textbf{Bolded} numbers indicate the best of proxy vs.\ lattice. \textsc{Speedup} is the relative speed increase for marginalization when using lattice sampling instead of importance sampling.}
\label{tab:translations}
\end{table}

We additionally wanted to experiment on a more free form task which would have longer outputs compared to Q\&A. We used the \textsc{Gemma3} model family to do zero-shot English$\rightarrow$German and English$\rightarrow$Czech translation. We used 150 sentences for each language pair from the \textsc{WMT24++} translation dataset \cite{deutsch2025wmt24expandinglanguagecoverage} for the datasets.

We first used the model to generate $k = 4$ unique translations of the input sentence. We then computed the canonical tokenization probability and the non-canonical marginal for each translation and chose the highest one for our output. The resulting translated corpus was scored with \textsc{Comet}\textsubscript{22}.

We find that all three sampling strategies perform well, with no clear winner among them across model sizes or language pairs. However, like Q\&A, lattice sampling was significantly faster than importance sampling.

This again supports our main claims: non-canonical marginalization encodes a useful signal and our marginal approximation method provides more accurate marginal estimates than proxy sampling at a much lower runtime cost.

\subsection{Intrinsic Experiment: Proxy Importance Sampling Underestimates the Marginal} \label{ssec:underestimation_experiment}

In Section \ref{ssec:enumeration}, we noted that our lattice sampler provides a strict lower-bound on the marginal, as it generates and scores sequences without replacement. We reanalyze the marginal estimates from both the Q\&A and translation datasets to determine how often the importance sampler estimated the marginal to be less than what the lattice sampler produced. Since they used the same number of samples, these cases would be where the importance sampler verifiably underestimated the marginal. Note that the lattice sampler also, by definition, underestimates the marginal but if it is higher than the importance sampler estimate, then it is guaranteed to be a better estimate of the marginal.

Concretely, for each example in a dataset and for each of its possible choices (all of the multiple choice options for Q\&A and the each of $k$ candidate translations for the translation tasks; not just the highest scoring ones), we compared the marginal estimates under each sampling scheme. Table \ref{tab:intrinsic_understimate} gives the results for each dataset.

In every dataset and model paring, proxy importance sampling estimated a value that was lower than the lattice sampler more than 50\% of the times. In the worse case, for \textsc{Arc} and the smaller \textsc{Gemma3} models, it was as bad as $82\%$ of cases being underestimated. This is despite the lattice sampler and the proxy importance sampler using the same number of samples.

Thus, we see that importance sampling consistently underestimates the true marginal compared to lattice sampling, meaning that lattice sampling produces estimates that are more accurate but also substantially faster to compute.

\begin{table}[]
\centering
\begin{tabular}{lccccc}
\toprule
\multirow{2}{*}{\textsc{Model}} & \multicolumn{5}{c}{Proxy Sampling  Underestimation Ratio}      \\
      & \textsc{Book} & \textsc{Med}       & \textsc{ARC} & En$\rightarrow$De & En$\rightarrow$Cs \\ \midrule
\textsc{Llama2-7b} & $72.3$    & $76.7$          & $74.2$ & - & -     \\ \midrule
\textsc{Llama2-13b} & $74.8$    & $69.1$          & $72.2$ & - & -     \\ \midrule
\textsc{Gemma3-1b} & $76.3$    & $73.3$          & $82.8$ & $52.2$ & $62.3$     \\ \midrule
\textsc{Gemma3-4b} & $76.0$    & $78.8$          & $82.8$ & $72.4$ & $58.3$     \\ \midrule
\textsc{Gemma3-27b} & $76.4$    & $78.5$          & $78.1$ & $72.4$ & $74.0$     \\ \midrule
\end{tabular}
\caption{The percentage of examples (answers for Q\&A and candidate translations) where proxy importance sampling underestimated the true marginal. This is measured by checking if importance sampling's estimate was less than lattice sampling's estimate, which is guaranteed to be a lower-bound on the true marginal. \textbf{NB:} We only used \textsc{Gemma3} for translation.}\label{tab:intrinsic_understimate}
\end{table}

\section{Conclusion}
Motivated by approximating the marginal distribution for LLMs but avoiding the expensive LLM generation step required by importance sampling, we presented a \textit{decoding-free} approximate LLM marginalization algorithm based on the tokenization lattice --- the automaton that succinctly encodes all possible tokenizations of an input text.  Our method is tokenizer-agnostic, extremely fast, and provides provable lower bounds on the marginal via sampling without replacement. We entirely side-step the generation process and only require the LLM to \textit{score} each sampled tokenization, which is comparatively cheap.

On two downstream tasks, we demonstrate that non-canonical marginalization is a viable inference strategy compared to only scoring the canonical tokenization of a sequence. Additionally, we find that our lattice sampler matches or beats the expensive LLM importance sampler using the same number of samples but with large runtime gains (up to $>\!30x$ speedups). Finally, we show that the proxy importance sampler verifiably underestimates the true marginal in the majority of cases in our downstream tasks.

Thus, our method provides a cheap and powerful alternative to LLM importance sampling for marginalization.

\section*{Limitations}

Our primary limitations were nearly all due to the large computational cost involved in approximating marginals with LLMs via importance sampling. For example, we used smaller subsets of datasets and did not experiment with very large ($>\!$70b parameter) LLMs or techniques like prompt-tuning, temperature tuning, or speculative decoding all due to the prohibitively large computational costs involved. However, our goal is not to demonstrate state-of-the-art performance, but rather to show that our method is comparable with importance sampling, but faster.

We also only dealt with open source LLMs, as we required direct logit access in order to produce the constrained LLM for importance sampling.

\bibliography{aaai2026}

\begin{thebibliography}{33}
\providecommand{\natexlab}[1]{#1}

\bibitem[{Allauzen and Mohri(2009)}]{allauzen-mohri-2009}
Allauzen, C.; and Mohri, M. 2009.
\newblock N-Way Composition of Weighted Finite-State Transducers.
\newblock \emph{International Journal of Foundations of Computer Science}, 20(04): 613--627.

\bibitem[{Bauwens, Kacz{\'e}r, and De~Lhoneux(2025)}]{bauwens-etal-2025-grampa}
Bauwens, T.; Kacz{\'e}r, D.; and De~Lhoneux, M. 2025.
\newblock {GR}a{MP}a: Subword Regularisation by Skewing Uniform Segmentation Distributions with an Efficient Path-counting {M}arkov Model.
\newblock In Che, W.; Nabende, J.; Shutova, E.; and Pilehvar, M.~T., eds., \emph{Proceedings of the 63rd Annual Meeting of the Association for Computational Linguistics (Volume 1: Long Papers)}, 24228--24257. Vienna, Austria: Association for Computational Linguistics.
\newblock ISBN 979-8-89176-251-0.

\bibitem[{Brown et~al.(2020)Brown, Mann, Ryder, Subbiah, Kaplan, Dhariwal, Neelakantan, Shyam, Sastry, Askell, Agarwal, Herbert{-}Voss, Krueger, Henighan, Child, Ramesh, Ziegler, Wu, Winter, Hesse, Chen, Sigler, Litwin, Gray, Chess, Clark, Berner, McCandlish, Radford, Sutskever, and Amodei}]{gpt}
Brown, T.~B.; Mann, B.; Ryder, N.; Subbiah, M.; Kaplan, J.; Dhariwal, P.; Neelakantan, A.; Shyam, P.; Sastry, G.; Askell, A.; Agarwal, S.; Herbert{-}Voss, A.; Krueger, G.; Henighan, T.; Child, R.; Ramesh, A.; Ziegler, D.~M.; Wu, J.; Winter, C.; Hesse, C.; Chen, M.; Sigler, E.; Litwin, M.; Gray, S.; Chess, B.; Clark, J.; Berner, C.; McCandlish, S.; Radford, A.; Sutskever, I.; and Amodei, D. 2020.
\newblock Language Models are Few-Shot Learners.
\newblock In \emph{Advances in Neural Information Processing Systems 33: Annual Conference on Neural Information Processing Systems 2020, NeurIPS 2020, December 6-12, 2020, virtual}.

\bibitem[{Cao and Rimell(2021)}]{cao-rimell-2021-evaluate}
Cao, K.; and Rimell, L. 2021.
\newblock You should evaluate your language model on marginal likelihood over tokenisations.
\newblock In Moens, M.-F.; Huang, X.; Specia, L.; and Yih, S. W.-t., eds., \emph{Proceedings of the 2021 Conference on Empirical Methods in Natural Language Processing}, 2104--2114. Online and Punta Cana, Dominican Republic: Association for Computational Linguistics.

\bibitem[{Chirkova et~al.(2023)Chirkova, Kruszewski, Rozen, and Dymetman}]{chirkova-etal-2023-marginalize}
Chirkova, N.; Kruszewski, G.; Rozen, J.; and Dymetman, M. 2023.
\newblock Should you marginalize over possible tokenizations?
\newblock In Rogers, A.; Boyd-Graber, J.; and Okazaki, N., eds., \emph{Proceedings of the 61st Annual Meeting of the Association for Computational Linguistics (Volume 2: Short Papers)}, 1--12. Toronto, Canada: Association for Computational Linguistics.

\bibitem[{Clark et~al.(2018)Clark, Cowhey, Etzioni, Khot, Sabharwal, Schoenick, and Tafjord}]{allenai:arc}
Clark, P.; Cowhey, I.; Etzioni, O.; Khot, T.; Sabharwal, A.; Schoenick, C.; and Tafjord, O. 2018.
\newblock Think you have Solved Question Answering? Try ARC, the AI2 Reasoning Challenge.
\newblock \emph{arXiv:1803.05457v1}.

\bibitem[{Cognetta and Okazaki(2024)}]{cognetta2024tokenizationfinitestatetransduction}
Cognetta, M.; and Okazaki, N. 2024.
\newblock Tokenization as Finite-State Transduction.
\newblock arXiv:2410.15696.

\bibitem[{Cognetta, Zouhar, and Okazaki(2024)}]{cognetta-etal-2024-distributional}
Cognetta, M.; Zouhar, V.; and Okazaki, N. 2024.
\newblock Distributional Properties of Subword Regularization.
\newblock In Al-Onaizan, Y.; Bansal, M.; and Chen, Y.-N., eds., \emph{Proceedings of the 2024 Conference on Empirical Methods in Natural Language Processing}, 10753--10763. Miami, Florida, USA: Association for Computational Linguistics.

\bibitem[{Deutsch et~al.(2025)Deutsch, Briakou, Caswell, Finkelstein, Galor, Juraska, Kovacs, Lui, Rei, Riesa, Rijhwani, Riley, Salesky, Trabelsi, Winkler, Zhang, and Freitag}]{deutsch2025wmt24expandinglanguagecoverage}
Deutsch, D.; Briakou, E.; Caswell, I.; Finkelstein, M.; Galor, R.; Juraska, J.; Kovacs, G.; Lui, A.; Rei, R.; Riesa, J.; Rijhwani, S.; Riley, P.; Salesky, E.; Trabelsi, F.; Winkler, S.; Zhang, B.; and Freitag, M. 2025.
\newblock {WMT24++: Expanding the Language Coverage of WMT24 to 55 Languages \& Dialects}.
\newblock arXiv:2502.12404.

\bibitem[{Edman, Schmid, and Fraser(2024)}]{edman-etal-2024-cute}
Edman, L.; Schmid, H.; and Fraser, A. 2024.
\newblock {CUTE}: Measuring {LLM}s' Understanding of Their Tokens.
\newblock In Al-Onaizan, Y.; Bansal, M.; and Chen, Y.-N., eds., \emph{Proceedings of the 2024 Conference on Empirical Methods in Natural Language Processing}, 3017--3026. Miami, Florida, USA: Association for Computational Linguistics.

\bibitem[{Gall{\'e}(2019)}]{galle-2019-investigating}
Gall{\'e}, M. 2019.
\newblock Investigating the Effectiveness of {BPE}: The Power of Shorter Sequences.
\newblock In Inui, K.; Jiang, J.; Ng, V.; and Wan, X., eds., \emph{Proceedings of the 2019 Conference on Empirical Methods in Natural Language Processing and the 9th International Joint Conference on Natural Language Processing (EMNLP-IJCNLP)}, 1375--1381. Hong Kong, China: Association for Computational Linguistics.

\bibitem[{Geh et~al.(2024)Geh, Zhang, Ahmed, Wang, and Van Den~Broeck}]{geh-etal-2024-signal}
Geh, R.; Zhang, H.; Ahmed, K.; Wang, B.; and Van Den~Broeck, G. 2024.
\newblock Where is the signal in tokenization space?
\newblock In Al-Onaizan, Y.; Bansal, M.; and Chen, Y.-N., eds., \emph{Proceedings of the 2024 Conference on Empirical Methods in Natural Language Processing}, 3966--3979. Miami, Florida, USA: Association for Computational Linguistics.

\bibitem[{{Gemma Team} et~al.(2025){Gemma Team}, Kamath, Ferret, Pathak, Vieillard, Merhej, Perrin, Matejovicova, Ramé, Rivière, Rouillard, Mesnard, Cideron, bastien Grill, Ramos, Yvinec, Casbon, Pot, Penchev, Liu, Visin, Kenealy, Beyer, Zhai, Tsitsulin, Busa-Fekete, Feng, Sachdeva, Coleman, Gao, Mustafa, Barr, Parisotto, Tian, Eyal, Cherry, Peter, Sinopalnikov, Bhupatiraju, Agarwal, Kazemi, Malkin, Kumar, Vilar, Brusilovsky, Luo, Steiner, Friesen, Sharma, Sharma, Gilady, Goedeckemeyer, Saade, Feng, Kolesnikov, Bendebury, Abdagic, Vadi, György, Pinto, Das, Bapna, Miech, Yang, Paterson, Shenoy, Chakrabarti, Piot, Wu, Shahriari, Petrini, Chen, Lan, Choquette-Choo, Carey, Brick, Deutsch, Eisenbud, Cattle, Cheng, Paparas, Sreepathihalli, Reid, Tran, Zelle, Noland, Huizenga, Kharitonov, Liu, Amirkhanyan, Cameron, Hashemi, Klimczak-Plucińska, Singh, Mehta, Lehri, Hazimeh, Ballantyne, Szpektor, Nardini, Pouget-Abadie, Chan, Stanton, Wieting, Lai, Orbay, Fernandez, Newlan, yeong Ji, Singh, Black, Yu, Hui,
  Vodrahalli, Greff, Qiu, Valentine, Coelho, Ritter, Hoffman, Watson, Chaturvedi, Moynihan, Ma, Babar, Noy, Byrd, Roy, Momchev, Chauhan, Sachdeva, Bunyan, Botarda, Caron, Rubenstein, Culliton, Schmid, Sessa, Xu, Stanczyk, Tafti, Shivanna, Wu, Pan, Rokni, Willoughby, Vallu, Mullins, Jerome, Smoot, Girgin, Iqbal, Reddy, Sheth, Põder, Bhatnagar, Panyam, Eiger, Zhang, Liu, Yacovone, Liechty, Kalra, Evci, Misra, Roseberry, Feinberg, Kolesnikov, Han, Kwon, Chen, Chow, Zhu, Wei, Egyed, Cotruta, Giang, Kirk, Rao, Black, Babar, Lo, Moreira, Martins, Sanseviero, Gonzalez, Gleicher, Warkentin, Mirrokni, Senter, Collins, Barral, Ghahramani, Hadsell, Matias, Sculley, Petrov, Fiedel, Shazeer, Vinyals, Dean, Hassabis, Kavukcuoglu, Farabet, Buchatskaya, Alayrac, Anil, Dmitry, Lepikhin, Borgeaud, Bachem, Joulin, Andreev, Hardin, Dadashi, and Hussenot}]{gemma3}
{Gemma Team}; Kamath, A.; Ferret, J.; Pathak, S.; Vieillard, N.; Merhej, R.; Perrin, S.; Matejovicova, T.; Ramé, A.; Rivière, M.; Rouillard, L.; Mesnard, T.; Cideron, G.; bastien Grill, J.; Ramos, S.; Yvinec, E.; Casbon, M.; Pot, E.; Penchev, I.; Liu, G.; Visin, F.; Kenealy, K.; Beyer, L.; Zhai, X.; Tsitsulin, A.; Busa-Fekete, R.; Feng, A.; Sachdeva, N.; Coleman, B.; Gao, Y.; Mustafa, B.; Barr, I.; Parisotto, E.; Tian, D.; Eyal, M.; Cherry, C.; Peter, J.-T.; Sinopalnikov, D.; Bhupatiraju, S.; Agarwal, R.; Kazemi, M.; Malkin, D.; Kumar, R.; Vilar, D.; Brusilovsky, I.; Luo, J.; Steiner, A.; Friesen, A.; Sharma, A.; Sharma, A.; Gilady, A.~M.; Goedeckemeyer, A.; Saade, A.; Feng, A.; Kolesnikov, A.; Bendebury, A.; Abdagic, A.; Vadi, A.; György, A.; Pinto, A.~S.; Das, A.; Bapna, A.; Miech, A.; Yang, A.; Paterson, A.; Shenoy, A.; Chakrabarti, A.; Piot, B.; Wu, B.; Shahriari, B.; Petrini, B.; Chen, C.; Lan, C.~L.; Choquette-Choo, C.~A.; Carey, C.; Brick, C.; Deutsch, D.; Eisenbud, D.; Cattle, D.; Cheng, D.;
  Paparas, D.; Sreepathihalli, D.~S.; Reid, D.; Tran, D.; Zelle, D.; Noland, E.; Huizenga, E.; Kharitonov, E.; Liu, F.; Amirkhanyan, G.; Cameron, G.; Hashemi, H.; Klimczak-Plucińska, H.; Singh, H.; Mehta, H.; Lehri, H.~T.; Hazimeh, H.; Ballantyne, I.; Szpektor, I.; Nardini, I.; Pouget-Abadie, J.; Chan, J.; Stanton, J.; Wieting, J.; Lai, J.; Orbay, J.; Fernandez, J.; Newlan, J.; yeong Ji, J.; Singh, J.; Black, K.; Yu, K.; Hui, K.; Vodrahalli, K.; Greff, K.; Qiu, L.; Valentine, M.; Coelho, M.; Ritter, M.; Hoffman, M.; Watson, M.; Chaturvedi, M.; Moynihan, M.; Ma, M.; Babar, N.; Noy, N.; Byrd, N.; Roy, N.; Momchev, N.; Chauhan, N.; Sachdeva, N.; Bunyan, O.; Botarda, P.; Caron, P.; Rubenstein, P.~K.; Culliton, P.; Schmid, P.; Sessa, P.~G.; Xu, P.; Stanczyk, P.; Tafti, P.; Shivanna, R.; Wu, R.; Pan, R.; Rokni, R.; Willoughby, R.; Vallu, R.; Mullins, R.; Jerome, S.; Smoot, S.; Girgin, S.; Iqbal, S.; Reddy, S.; Sheth, S.; Põder, S.; Bhatnagar, S.; Panyam, S.~R.; Eiger, S.; Zhang, S.; Liu, T.; Yacovone, T.;
  Liechty, T.; Kalra, U.; Evci, U.; Misra, V.; Roseberry, V.; Feinberg, V.; Kolesnikov, V.; Han, W.; Kwon, W.; Chen, X.; Chow, Y.; Zhu, Y.; Wei, Z.; Egyed, Z.; Cotruta, V.; Giang, M.; Kirk, P.; Rao, A.; Black, K.; Babar, N.; Lo, J.; Moreira, E.; Martins, L.~G.; Sanseviero, O.; Gonzalez, L.; Gleicher, Z.; Warkentin, T.; Mirrokni, V.; Senter, E.; Collins, E.; Barral, J.; Ghahramani, Z.; Hadsell, R.; Matias, Y.; Sculley, D.; Petrov, S.; Fiedel, N.; Shazeer, N.; Vinyals, O.; Dean, J.; Hassabis, D.; Kavukcuoglu, K.; Farabet, C.; Buchatskaya, E.; Alayrac, J.-B.; Anil, R.; Dmitry; Lepikhin; Borgeaud, S.; Bachem, O.; Joulin, A.; Andreev, A.; Hardin, C.; Dadashi, R.; and Hussenot, L. 2025.
\newblock Gemma 3 Technical Report.
\newblock arXiv:2503.19786.

\bibitem[{Hiraoka(2022)}]{hiraoka-2022-maxmatch}
Hiraoka, T. 2022.
\newblock {M}ax{M}atch-Dropout: Subword Regularization for {W}ord{P}iece.
\newblock In Calzolari, N.; Huang, C.-R.; Kim, H.; Pustejovsky, J.; Wanner, L.; Choi, K.-S.; Ryu, P.-M.; Chen, H.-H.; Donatelli, L.; Ji, H.; Kurohashi, S.; Paggio, P.; Xue, N.; Kim, S.; Hahm, Y.; He, Z.; Lee, T.~K.; Santus, E.; Bond, F.; and Na, S.-H., eds., \emph{Proceedings of the 29th International Conference on Computational Linguistics}, 4864--4872. Gyeongju, Republic of Korea: International Committee on Computational Linguistics.

\bibitem[{Hiraoka and Okazaki(2024)}]{DBLP:journals/corr/abs-2402-09808}
Hiraoka, T.; and Okazaki, N. 2024.
\newblock Knowledge of Pretrained Language Models on Surface Information of Tokens.
\newblock \emph{CoRR}, abs/2402.09808.

\bibitem[{Hiraoka, Shindo, and Matsumoto(2019)}]{hiraoka-etal-2019-stochastic}
Hiraoka, T.; Shindo, H.; and Matsumoto, Y. 2019.
\newblock Stochastic Tokenization with a Language Model for Neural Text Classification.
\newblock In Korhonen, A.; Traum, D.; and M{\`a}rquez, L., eds., \emph{Proceedings of the 57th Annual Meeting of the Association for Computational Linguistics}, 1620--1629. Florence, Italy: Association for Computational Linguistics.

\bibitem[{Hoens(2008)}]{hoens2008counting}
Hoens, T.~R. 2008.
\newblock \emph{Counting and sampling paths in graphs}.
\newblock M.s. thesis, Rochester Institute of Technology, Rochester, New York.

\bibitem[{Kaushal and Mahowald(2022)}]{kaushal-mahowald-2022-tokens}
Kaushal, A.; and Mahowald, K. 2022.
\newblock What do tokens know about their characters and how do they know it?
\newblock In Carpuat, M.; de~Marneffe, M.-C.; and Meza~Ruiz, I.~V., eds., \emph{Proceedings of the 2022 Conference of the North American Chapter of the Association for Computational Linguistics: Human Language Technologies}, 2487--2507. Seattle, United States: Association for Computational Linguistics.

\bibitem[{Koo, Liu, and He(2024)}]{koo}
Koo, T.; Liu, F.; and He, L. 2024.
\newblock Automata-based constraints for language model decoding.
\newblock In \emph{First Conference on Language Modeling}.

\bibitem[{Kudo(2018)}]{kudo-2018-subword}
Kudo, T. 2018.
\newblock Subword Regularization: Improving Neural Network Translation Models with Multiple Subword Candidates.
\newblock In Gurevych, I.; and Miyao, Y., eds., \emph{Proceedings of the 56th Annual Meeting of the Association for Computational Linguistics (Volume 1: Long Papers)}, 66--75. Melbourne, Australia: Association for Computational Linguistics.

\bibitem[{Mielke et~al.(2021)Mielke, Alyafeai, Salesky, Raffel, Dey, Gallé, Raja, Si, Lee, Sagot, and Tan}]{mielke2021wordscharactersbriefhistory}
Mielke, S.~J.; Alyafeai, Z.; Salesky, E.; Raffel, C.; Dey, M.; Gallé, M.; Raja, A.; Si, C.; Lee, W.~Y.; Sagot, B.; and Tan, S. 2021.
\newblock Between words and characters: A Brief History of Open-Vocabulary Modeling and Tokenization in NLP.
\newblock arXiv:2112.10508.

\bibitem[{Mihaylov et~al.(2018)Mihaylov, Clark, Khot, and Sabharwal}]{OpenBookQA2018}
Mihaylov, T.; Clark, P.; Khot, T.; and Sabharwal, A. 2018.
\newblock Can a Suit of Armor Conduct Electricity? A New Dataset for Open Book Question Answering.
\newblock In \emph{EMNLP}.

\bibitem[{Murray and Chiang(2018)}]{murray-chiang-2018-correcting}
Murray, K.; and Chiang, D. 2018.
\newblock Correcting Length Bias in Neural Machine Translation.
\newblock In Bojar, O.; Chatterjee, R.; Federmann, C.; Fishel, M.; Graham, Y.; Haddow, B.; Huck, M.; Yepes, A.~J.; Koehn, P.; Monz, C.; Negri, M.; N{\'e}v{\'e}ol, A.; Neves, M.; Post, M.; Specia, L.; Turchi, M.; and Verspoor, K., eds., \emph{Proceedings of the Third Conference on Machine Translation: Research Papers}, 212--223. Brussels, Belgium: Association for Computational Linguistics.

\bibitem[{Pal, Umapathi, and Sankarasubbu(2022)}]{pmlr-v174-pal22a}
Pal, A.; Umapathi, L.~K.; and Sankarasubbu, M. 2022.
\newblock MedMCQA: A Large-scale Multi-Subject Multi-Choice Dataset for Medical domain Question Answering.
\newblock In Flores, G.; Chen, G.~H.; Pollard, T.; Ho, J.~C.; and Naumann, T., eds., \emph{Proceedings of the Conference on Health, Inference, and Learning}, volume 174 of \emph{Proceedings of Machine Learning Research}, 248--260. PMLR.

\bibitem[{Provilkov, Emelianenko, and Voita(2020)}]{provilkov-etal-2020-bpe}
Provilkov, I.; Emelianenko, D.; and Voita, E. 2020.
\newblock {BPE}-Dropout: Simple and Effective Subword Regularization.
\newblock In Jurafsky, D.; Chai, J.; Schluter, N.; and Tetreault, J., eds., \emph{Proceedings of the 58th Annual Meeting of the Association for Computational Linguistics}, 1882--1892. Online: Association for Computational Linguistics.

\bibitem[{Qwen et~al.(2025)Qwen, :, Yang, Yang, Zhang, Hui, Zheng, Yu, Li, Liu, Huang, Wei, Lin, Yang, Tu, Zhang, Yang, Yang, Zhou, Lin, Dang, Lu, Bao, Yang, Yu, Li, Xue, Zhang, Zhu, Men, Lin, Li, Tang, Xia, Ren, Ren, Fan, Su, Zhang, Wan, Liu, Cui, Zhang, and Qiu}]{qwen}
Qwen; :; Yang, A.; Yang, B.; Zhang, B.; Hui, B.; Zheng, B.; Yu, B.; Li, C.; Liu, D.; Huang, F.; Wei, H.; Lin, H.; Yang, J.; Tu, J.; Zhang, J.; Yang, J.; Yang, J.; Zhou, J.; Lin, J.; Dang, K.; Lu, K.; Bao, K.; Yang, K.; Yu, L.; Li, M.; Xue, M.; Zhang, P.; Zhu, Q.; Men, R.; Lin, R.; Li, T.; Tang, T.; Xia, T.; Ren, X.; Ren, X.; Fan, Y.; Su, Y.; Zhang, Y.; Wan, Y.; Liu, Y.; Cui, Z.; Zhang, Z.; and Qiu, Z. 2025.
\newblock Qwen2.5 Technical Report.
\newblock arXiv:2412.15115.

\bibitem[{Sennrich, Haddow, and Birch(2016)}]{sennrich-etal-2016-neural}
Sennrich, R.; Haddow, B.; and Birch, A. 2016.
\newblock Neural Machine Translation of Rare Words with Subword Units.
\newblock In Erk, K.; and Smith, N.~A., eds., \emph{Proceedings of the 54th Annual Meeting of the Association for Computational Linguistics (Volume 1: Long Papers)}, 1715--1725. Berlin, Germany: Association for Computational Linguistics.

\bibitem[{Touvron et~al.(2023)Touvron, Martin, Stone, Albert, Almahairi, Babaei, Bashlykov, Batra, Bhargava, Bhosale, Bikel, Blecher, Ferrer, Chen, Cucurull, Esiobu, Fernandes, Fu, Fu, Fuller, Gao, Goswami, Goyal, Hartshorn, Hosseini, Hou, Inan, Kardas, Kerkez, Khabsa, Kloumann, Korenev, Koura, Lachaux, Lavril, Lee, Liskovich, Lu, Mao, Martinet, Mihaylov, Mishra, Molybog, Nie, Poulton, Reizenstein, Rungta, Saladi, Schelten, Silva, Smith, Subramanian, Tan, Tang, Taylor, Williams, Kuan, Xu, Yan, Zarov, Zhang, Fan, Kambadur, Narang, Rodriguez, Stojnic, Edunov, and Scialom}]{llama}
Touvron, H.; Martin, L.; Stone, K.; Albert, P.; Almahairi, A.; Babaei, Y.; Bashlykov, N.; Batra, S.; Bhargava, P.; Bhosale, S.; Bikel, D.; Blecher, L.; Ferrer, C.~C.; Chen, M.; Cucurull, G.; Esiobu, D.; Fernandes, J.; Fu, J.; Fu, W.; Fuller, B.; Gao, C.; Goswami, V.; Goyal, N.; Hartshorn, A.; Hosseini, S.; Hou, R.; Inan, H.; Kardas, M.; Kerkez, V.; Khabsa, M.; Kloumann, I.; Korenev, A.; Koura, P.~S.; Lachaux, M.-A.; Lavril, T.; Lee, J.; Liskovich, D.; Lu, Y.; Mao, Y.; Martinet, X.; Mihaylov, T.; Mishra, P.; Molybog, I.; Nie, Y.; Poulton, A.; Reizenstein, J.; Rungta, R.; Saladi, K.; Schelten, A.; Silva, R.; Smith, E.~M.; Subramanian, R.; Tan, X.~E.; Tang, B.; Taylor, R.; Williams, A.; Kuan, J.~X.; Xu, P.; Yan, Z.; Zarov, I.; Zhang, Y.; Fan, A.; Kambadur, M.; Narang, S.; Rodriguez, A.; Stojnic, R.; Edunov, S.; and Scialom, T. 2023.
\newblock Llama 2: Open Foundation and Fine-Tuned Chat Models.
\newblock arXiv:2307.09288.

\bibitem[{Vieira et~al.(2025)Vieira, Liu, Pasti, Emara, DuSell, LeBrun, Giulianelli, Gastaldi, Terilla, O'Donnell, and Cotterell}]{vieira2025languagemodelscanonicalbytepair}
Vieira, T.; Liu, T.; Pasti, C.; Emara, Y.; DuSell, B.; LeBrun, B.; Giulianelli, M.; Gastaldi, J.~L.; Terilla, J.; O'Donnell, T.~J.; and Cotterell, R. 2025.
\newblock Language Models over Canonical Byte-Pair Encodings.
\newblock In \emph{Forty-second International Conference on Machine Learning}.

\bibitem[{Welleck et~al.(2024)Welleck, Bertsch, Finlayson, Schoelkopf, Xie, Neubig, Kulikov, and Harchaoui}]{DBLP:journals/tmlr/WelleckBFSXNKH24}
Welleck, S.; Bertsch, A.; Finlayson, M.; Schoelkopf, H.; Xie, A.; Neubig, G.; Kulikov, I.; and Harchaoui, Z. 2024.
\newblock From Decoding to Meta-Generation: Inference-time Algorithms for Large Language Models.
\newblock \emph{Trans. Mach. Learn. Res.}, 2024.

\bibitem[{Willard and Louf(2023)}]{willard2023efficientguidedgenerationlarge}
Willard, B.~T.; and Louf, R. 2023.
\newblock Efficient Guided Generation for Large Language Models.
\newblock arXiv:2307.09702.

\bibitem[{Wolf et~al.(2020)Wolf, Debut, Sanh, Chaumond, Delangue, Moi, Cistac, Rault, Louf, Funtowicz, Davison, Shleifer, von Platen, Ma, Jernite, Plu, Xu, Scao, Gugger, Drame, Lhoest, and Rush}]{wolf-etal-2020-transformers}
Wolf, T.; Debut, L.; Sanh, V.; Chaumond, J.; Delangue, C.; Moi, A.; Cistac, P.; Rault, T.; Louf, R.; Funtowicz, M.; Davison, J.; Shleifer, S.; von Platen, P.; Ma, C.; Jernite, Y.; Plu, J.; Xu, C.; Scao, T.~L.; Gugger, S.; Drame, M.; Lhoest, Q.; and Rush, A.~M. 2020.
\newblock Transformers: State-of-the-Art Natural Language Processing.
\newblock In \emph{Proceedings of the 2020 Conference on Empirical Methods in Natural Language Processing: System Demonstrations}, 38--45. Online: Association for Computational Linguistics.

\bibitem[{Zheng et~al.(2025)Zheng, Liu, Ahia, Hayase, Choi, and Smith}]{broken-tokens}
Zheng, B.~S.; Liu, A.; Ahia, O.; Hayase, J.; Choi, Y.; and Smith, N.~A. 2025.
\newblock Broken Tokens? Your Language Model can Secretly Handle Non-Canonical Tokenizations.
\newblock In \emph{Tokenization Workshop}.

\end{thebibliography}

\end{document}